\documentclass{article}
\usepackage{spconf,amsmath,graphicx}
\usepackage{xcolor}
\usepackage{hyperref}
\usepackage{array, booktabs, multirow}

\title{Full segmentation annotations\\ of 3D time-lapse microscopy images of MDA231 cells}
\name{Aleksandra Melnikova \qquad Petr Matula\thanks{The work was funded by the Czech Science Foundation, project no.~GA21-20374S. The authors thank Martin Bu\v{c}o for the help with the annotations and the Cell Tracking Challenge organizers for providing annotation tools.}}

\address{Center for Biomedical Image Analysis (CBIA), Faculty of Informatics,\\ Masaryk University, Brno, Czech Republic}
\begin{document}
%
\maketitle
\begin{abstract}

High-quality, publicly available segmentation annotations of image and video datasets are critical for advancing the field of image processing. In particular, annotations of volumetric images of a large number of targets are time-consuming and challenging. In \cite{melnikova2025study}, we presented the first publicly available full 3D time-lapse segmentation annotations of migrating cells with complex dynamic shapes. Concretely, three distinct humans annotated two sequences of MDA231 human breast carcinoma cells (Fluo-C3DL-MDA231) from the Cell Tracking Challenge (CTC).

This paper aims to provide a comprehensive description of the dataset and accompanying experiments that were not included in \cite{melnikova2025study} due to limitations in publication space. Namely, we show that the created annotations are consistent with the previously published tracking markers provided by the CTC organizers and the segmentation accuracy measured based on the 2D gold truth of CTC is within the inter-annotator variability margins. We compared the created 3D annotations with automatically created silver truth provided by CTC. We have found the proposed annotations better represent the complexity of the input images. The presented annotations can be used for testing and training cell segmentation, or analyzing 3D shapes of highly dynamic objects. 
\end{abstract}

\begin{keywords}
Annotated dataset, cell detection, cell segmentation, 3D, time-lapse annotations
\end{keywords}

\section{Introduction}
\label{sec:intro}

Annotated datasets play a pivotal role in developing, testing, and validating various image-processing algorithms. The deep learning methods, in particular, heavily rely on the existence of the annotated data for training. A rich group of datasets is needed for benchmarking algorithms of the same category, e.g., image segmentation and tracking methods. 

In the field of biomedical image analysis, one of the richest benchmark datasets for objective comparison of cell segmentation and tracking methods is the Cell Tracking Challenge (CTC)~\cite{mavska2014benchmark,ulman2017objective, mavska2023cell}. The CTC offers a wide variety of annotated datasets of microscopy images with different modalities and cell shapes.  The time-lapse nature of datasets allows training cell tracking algorithms along with segmentation algorithms. 

\begin{figure*}[htb]

\begin{minipage}[b]{1.0\linewidth}
  \centering
  \centerline{\includegraphics[width=16.5cm]{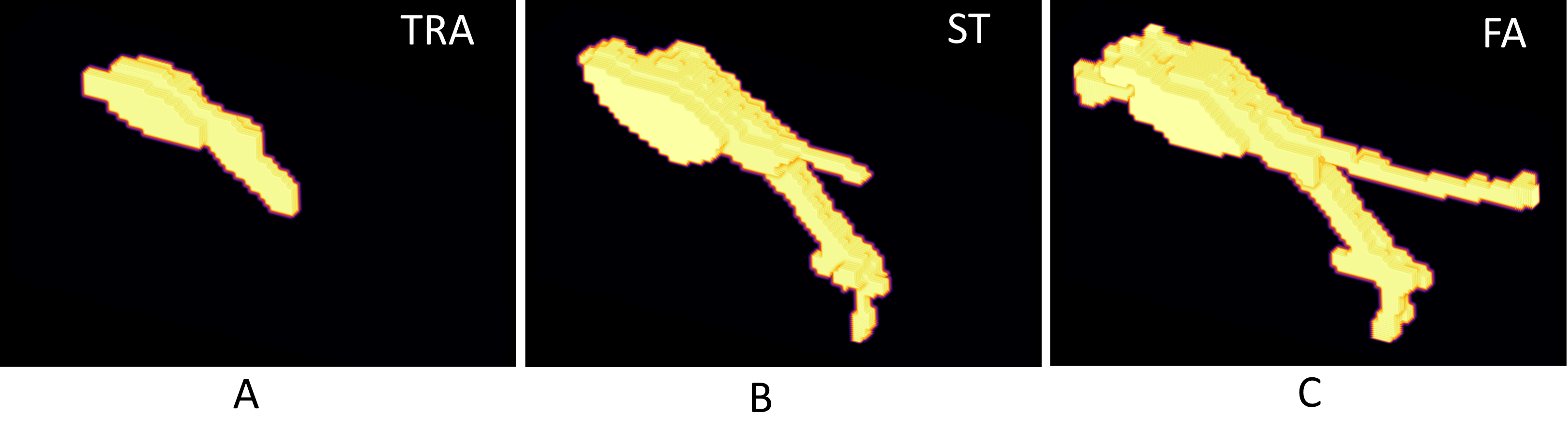}}
\end{minipage}
\begin{minipage}[b]{1.0\linewidth}
  \centering
  \centerline{\includegraphics[width=16.0cm]{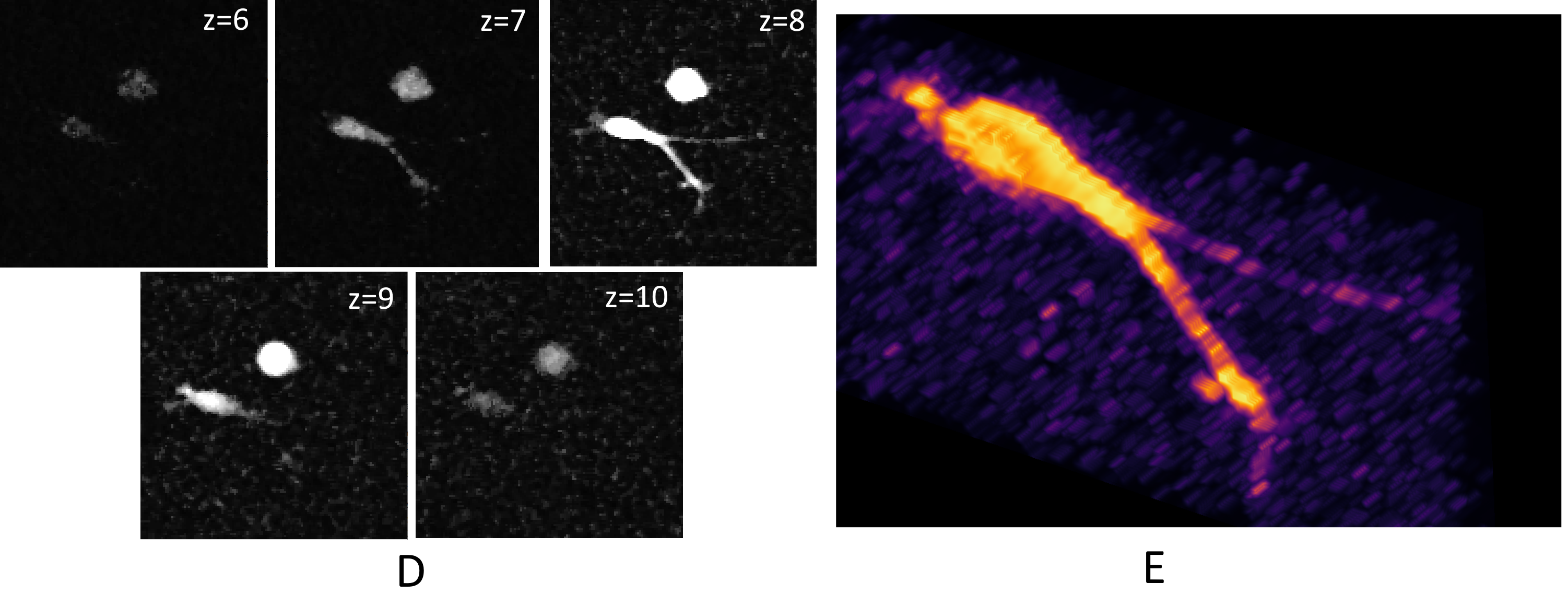}}
\end{minipage}
\hfill

\caption{ A 3D visualization of one cell mask that consists of 5 slices (w.r.t. the z-axis). The first row shows a 3D visualization of the 3 annotations: (A) Gold reference tracking annotations (TRA), (B) Silver reference segmentation annotations (ST), and (C) the proposed full annotations (FA). Image (D) shows all five 2D slices of the cell. Image (D) shows a 3D visualization of the slices (D). It is visible that thin protrusions are absent in images (A) and (B).}
\label{fig:vis}
\end{figure*}

The CTC offers three types of reference annotations, namely Gold reference tracking annotations (TRA), Gold reference segmentation annotations (GT), and Silver reference segmentation annotations (ST). The TRA fully covers all cell instances with small markers but does not provide full cell region information. The SEG is a collection of 2D binary masks of lateral cell sections for a subset of instances. The ST are computer-generated reference annotations obtained from several algorithms submitted by former challenge participants. See Figure \ref{fig:vis}A,B for an example of TRA and ST cell masks. No 3D CTC dataset has publicly available complete fully volumetric segmentation annotations.


In shape analysis, the shape can be used for the identification of common features or patterns in object morphology, e.g., for the analysis of cell nuclei \cite{bauer2017numerical}. The morphological features of cells can, for example, be a marker for pathology and relate to cell functions \cite{prasad2019cell, alizadeh2020cellular}. Therefore, it is important to extract cell shapes as precise as possible, including thin protrusions.  Unfortunately, high-quality annotations of highly dynamic shapes, such as migrating cells are missing, see Section~\ref{sec:rw}. Therefore, in \cite{melnikova2025study}, we presented a fully volumetric manual annotations for two temporal sequences of a 3D CTC dataset, Fluo-C3DL-MDA231, which is characterized by the most complex cell shapes out of all 3D CTC datasets. 

In this paper, as the main contribution, we provide a comprehensive description of
the dataset presented in \cite{melnikova2025study} and perform quantitative experiments that could not be included in \cite{melnikova2025study}. The cell shapes in the Fluo-C3DL-MDA231 dataset are irregular and may have several thin protrusions. To the best of our knowledge, there were no publicly available datasets of fully annotated 3D cells in time-lapse sequences before \cite{melnikova2025study}.
We collected annotations of three humans for the first temporal sequence and annotations of one person for the second temporal sequence. We also fused the three annotations of the first sequence by majority voting. All new annotations were compared to the existing annotations in the CTC and we have found that they are consistent with TRA, their accuracy is within inter-annotator variability as compared to GT, and have better coverage and quality than ST, see Figure \ref{fig:vis}C for an example. As we provide three human annotations of the first sequence, the annotations can also be used for the development and testing of the fusion methods that are often used for merging annotations of several experts.

\section{Related work}
\label{sec:rw}

Although various annotated 3D datasets are available for fields such as biology, medicine, robotics, or computer graphics, the interest in 3D image datasets has increased in recent years.  The 3D data is usually presented as one of the following types: meshes, point clouds, or voxels. The ShapeNet \cite{chang2015shapenet}, one of the richest 3D repositories for 3D shape classification, offers 3D polygonal models along with a rich set of annotations for each shape and their correspondences. The Stanford Joint 2D-3D-Semantic Data for Indoor Scene Understanding \cite{armeni2017joint} offers 2D images, 3D textured and semantic meshes, and 3D point cloud data of 6 large-scale indoor areas along with object class annotations. The overview of typically used 3D datasets for shape classification, object detection and tracking, and point cloud segmentation was presented in \cite{guo2020deep}. 3D datasets of natural images such as Waymo Open \cite{sun2020scalability}, a dataset of images for autonomous driving research, or Toronto-3D \cite{tan2020toronto}, a dataset of outdoor large-scale scenes, may contain hundreds or thousands of samples. Similarly, the medical field offers a wide variety of datasets, e.g., LIDC-IDRI \cite{armato2011lung}, a dataset of CT lung lesions containing 1018 CT scans or BraTS \cite{menze2014multimodal}, a dataset in the scope of Brain Tumor Segmentation Challenge. One of the richest repositories in the medical domain is NBIA-TCIA, The Cancer Imaging Archive \cite{clark2013cancer}, which uses National Biomedical Image Archive software. The datasets of microscopy images of cells in the biomedical field are different from these datasets as the images of living cells are typically highly dynamic, heterogeneous, and contain a large number of objects.

The diversity of microscopy image modalities and cell morphology, along with the presence of noise, makes the creation of annotated datasets challenging. In the biomedical field, by the annotation, the semantic annotation is usually implied. A widely known repository with public data is the BROAD Bioimage Benchmark Collection \cite{ljosa2012annotated}, which contains various annotated datasets of cells, and Image Data Resource \cite{Williams2017}, which is an open-source platform for publishing imaging data including related annotations. However, to our knowledge, the only fully annotated 3D datasets are synthetic \cite{maska2019toward, sorokin2018filogen, svoboda2011generation, svoboda2009generation}. Another well-known repository was created to objectively compare and benchmark cell segmentation and tracking methods: the Cell Tracking Challenge (CTC) \cite{mavska2014benchmark, mavska2023cell} and Time-Lapse Cell Segmentation Benchmark (CSB) became one of the world's richest repositories of annotated public datasets of cells. However, the 3D CTC datasets provide only manual segmentation annotations of random 2D slices.

For microscopy images, a large number of annotated objects is usually not available due to the laborious process of manual annotation and the requirement to have human experts perform the task. The images may contain hundreds of cells, which may be difficult to distinguish from one another and from non-cell objects due to noise that is often present in microscopy images. In the biomedical field, the common practice is collecting the annotations from several people and applying a fusion algorithm on the collected annotations such as majority-voting (MV) \cite{kittler1998combining}, SIMPLE \cite{langerak2010label}, or STAPLE \cite{warfield2004simultaneous}. The result of such fusion is usually called gold truth reference data. We follow this practice and try to fill the gap with lacking volumetric annotations of highly dynamic cells with complex shapes by annotating the Fluo-C3DL-MDA231 dataset freely available from the CTC website.

\section{Terminology}
\label{sec:Terminology}

\begin{figure*}[htb]
\begin{minipage}[b]{1.0\linewidth}
  \centering
  \centerline{\includegraphics[width=17.0cm]{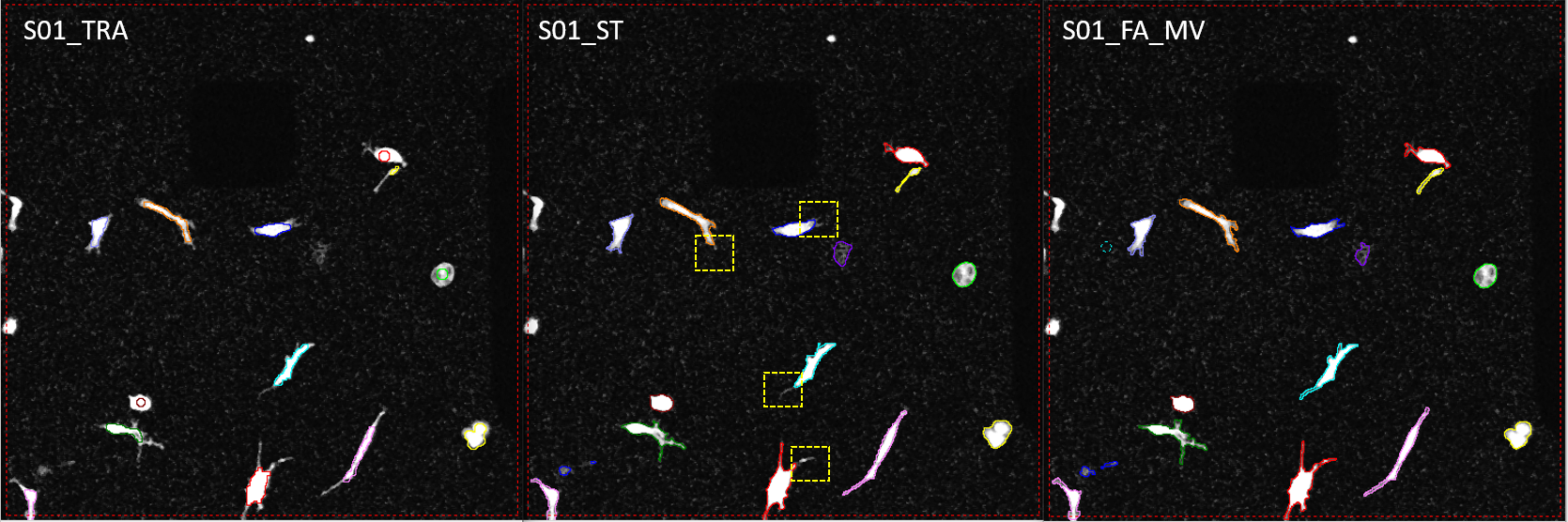}}
\end{minipage}
\caption{Overlay of 3 annotations (S01\_TRA and S01\_ST, and S01\_FA\_MV) over a raw image from the Fluo-C3DL-MDA231 dataset for a random 2D slice. The Figure shows that silver truth annotations have problems with the segmentation of thin protrusions (a yellow dashed bounding box denotes problematic regions).}
\label{fig:res}
\end{figure*}

We use CTC terminology and distinguish gold truth and silver truth reference annotations. The gold-standard corpus, or \textit{gold truth}, contains human-made reference annotations obtained as a result of the majority opinion of 3 human experts. The silver-standard corpus, or \textit{silver truth}, contains computer-generated reference annotations obtained as a result of majority voting over the results of several segmentation algorithms submitted by former challenge participants. We also distinguish \textit{segmentation} truth and \textit{tracking} truth. The former is used for training and evaluation of cell segmentation algorithms while the latter is used for training and validation of the cell tracking algorithms.

The CTC offers 4 sets of reference annotations for the Fluo-C3DL-MDA231 dataset: 1 set of gold tracking truth (TRA), random 2D slices of gold segmentation truth (GT), and 2 sets of silver segmentation truth. As the two silver truth datasets differ in labeling procedure only but the cell masks are the same, we consider them as one data set (ST) in this paper.

The raw images of the Fluo-C3DL-MDA231 dataset along with the gold truth and silver truth are publicly available without registration from the CTC website: \url{https://celltrackingchallenge.net/3d-datasets/}. The presented new annotations can be freely downloaded from CBIA website \url{https://cbia.fi.muni.cz/datasets/#isbi25dataset} or directly 
as\newline \url{https://datasets.gryf.fi.muni.cz/isbi2025/Fluo-C3DL-MDA231_Full_Annotations.zip}.

\section{Data collection}
\label{sec:Data_collection}

We provide annotations for the raw Fluo-C3DL-MDA231 dataset provided by CTC, which consists of two time-lapse sequences of images. The raw images were made by Dr. R. Kamm. Dept. of Biological Engineering, Massachusetts Institute of Technology, Cambridge MA (USA). Olympus FluoView F1000 microscope with an objective lens of Plan 20$\times$/0.7 microscope was used for acquisition. The images capture the MDA231 human breast carcinoma cells infected with a pMSCV vector, including the GFP sequence, embedded in a collagen matrix. The time step for both sequences was 80 minutes and twelve time points were captured. One voxel corresponds to $1.242 \times 1.242 \times 6.0$ microns. Each image has a size $30 \times 512 \times 512$ voxels.

All annotations were created manually with one of the two following procedures. Annotations made by annotator A1 were created by following protocol A. The annotations A2 and A3 were created by following protocol B. We used the same software that was used for collecting manual annotations for CTC \cite{mavska2023cell}.

\textit{Annotation protocol A:}
An annotator was asked to annotate all the cells present in all images. Then, cells were matched to the tracking markers (gold tracking truth) that were provided along with the raw images by CTC. The cells that were not matched to any marker were removed from the annotations.

\textit{Annotation protocol B:}
An annotator was asked to propagate or correct the tracking markers, i.e., only cells that had a tracking marker were annotated.

There are 3 annotations available (A1, A2, A3) for the first sequence (S01), and one annotation (A1) is available for the second sequence (S02). The annotations A1, A2, and A3 were merged by a simple majority-voting procedure to acquire the consensus annotation for sequence S01. The majority-voting procedure implies that all pixels having less than 2 votes for the same cell mask were removed from the final annotation. 
All annotations were saved as multi-layered \texttt{unit16} gray-scale \texttt{tif} images.

\section{Dataset Characteristics}
\label{sec:Dataset_overview}

In this section, we present the quantitative characteristics of the proposed annotations and compare them with all available sets of annotations provided by CTC. In Table~\ref{tab:all_stats}, we present the basic characteristics of all 3D datasets, namely, the gold tracking truth (TRA), silver truth (ST), and the proposed full annotations (FA). The table shows that the proposed annotations have a one-to-one relation to TRA, i.e., that they are perfectly consistent, where as ST has a slightly worse consistency with TRA. The table also shows that the bounding boxes of cells in FA are the largest, which is related to the fact that the protrusions are better annotated as can be seen in Figure~\ref{fig:res}.

\begin{table*}[htb]
\centering

\begin{tabular}{l|l|l|l|l|l|l|}
\toprule
\multirow{2}{*}{Property}
        &   \multicolumn{2}{c}{TRA (3D)}
                &   \multicolumn{2}{c}{ST (3D)} 
                        &   \multicolumn{2}{c}{FA (3D)} \\
    \cmidrule(lr){2-3}\cmidrule(lr){4-5}\cmidrule(lr){6-7}
                                  &  S01     &   S02     &   S01    &   S02    &   S01(MV)&   S02(A1)  \\
    \midrule
\# images                         &   12     &   12      &   12     &   12     &   12     &   12                      \\
\# masks                          &   \textbf{364}    &   \textbf{585}     &   351    &   582    &   \textbf{364}    &   \textbf{585}                    \\
\# masks per image                  &   \textbf{30.32}  &   \textbf{48.75}   &   29.25  &   48.5   &   \textbf{30.32}  &   \textbf{48.75}                    \\
Avg bounding box size X [voxels]  &   24.71  &   23.65   &   34.04  &   35.98  &   38.15  &   45.47 
\\
Avg bounding box size Y [voxels]  &   23.88  &   20.55   &   36.27  &   31.78  &   38.54  &   41.14 
\\
Avg bounding box size Z [voxels]  &   2.83   &   2.90    &   6.24   &   5.93   &   5.01    &   5.64 
\\
Avg volume per mask [voxels$^3$]    &   620.7 &   472.9  &   1781.2 &   1544.9 &   1496.4  &   2140.6 
\\
    \bottomrule
\end{tabular}
  \caption{The table shows a quantitative comparison of the characteristics between gold tracking truth (TRA), silver segmentation truth (ST), and proposed full annotations (FA). The measured characteristics consist of the number of images in sequence (\# images), the number of annotation masks in the whole sequence (\# masks), the average number of masks per one image (\# masks per image), average sizes of the 3D bounding box, and average volume per annotation mask. 
  }
\label{tab:all_stats}
\end{table*}

\subsection{Segmentation score}

The segmentation evaluation procedure of the CTC is based on computing the SEG score \cite{mavska2023cell}, which is a measure that calculates a mean Jaccard index $J$ between a reference $R$ and a segmentation $S$. The Jaccard index $J$ is computed for all matched pairs of masks ($r_i$, $s_i$) from $R$ and $S$. The pair is considered matched if the majority of the reference mask $r_i$ has an overlap with a segmentation instance $s_i$. If there is no mask $s_i$ that can be matched to $r_i$, $J_i$ is set to 0. The final segmentation score SEG equals the averaged sum of $J_i$ for all~$r_i$.

In Table~\ref{tab:gt_2d_comparison}, we present a quantitative comparison of the annotations with 2D gold truth annotations provided by CTC. The comparison was performed by measuring two measures: the SEG score and the average Hausdorff distance for objects that were matched by the SEG score matching procedure (i.e., if there is an overlap of more than $50\%$ with the gold truth mask). The results presented in the table show that the proposed annotations (FA) for sequence S01 are closer to the gold segmentation truth than ST. For sequence S02, the silver truth outperforms the proposed annotation due to the lack of annotations A2 and A3 for sequence S02.

\begin{table}[htb]
\centering
\begin{tabular}{l|l|l|l|l|l|l}
\toprule
\multirow{2}{*}{}
        &   \multicolumn{2}{c}{ST (3D)}
                &   \multicolumn{2}{c}{FA (3D)} 
                        \\
    \cmidrule(lr){2-3}\cmidrule(lr){4-5}
                  &   S01  &   S02     &   S01(MV) &  S02(A1)  \\
    \midrule
\#2d images       &   17   &   13      &   17      &  13                      \\
\#2d GT masks     &   121  &   103     &   121     &  103                      \\
\#matched masks   &   116  &   103     &   116     &  98                      \\

GT, SEG          &   0.7386 &  \textbf{0.751}   &   \textbf{0.752}   &   0.688              \\
GT, HD           &   5.11  &   \textbf{5.87}    &   \textbf{3.36}    &   6.54                 \\
                                      
    \bottomrule
\end{tabular}
  \caption{Quantitative comparison of the silver segmentation truth (ST) and the presented dataset of full annotations (FA) with the gold truth annotations (GT). The gold segmentation truth is available only for a limited number of random 2D slices (17 slices for the first sequence and 13 slices for the second sequence). The number of GT masks is also limited. In the table, a Jaccard index based measure (SEG) and averaged Hausdorff distance (HD) were computed between respective 2D slices for (GT, ST) and (GT, FA) pairs. See section \ref{sec:Dataset_overview} for more details on SEG measure and matching procedure.}
\label{tab:gt_2d_comparison}
\end{table}

\subsection{Inter-annotator variability}

The inter-annotator variability of the CTC GT (2D) annotations was calculated as the average of SEG scores between the GT and annotations of each annotator for both sequences, following the CTC practice. This characteristic shows the level of agreement between the annotators. In this subsection, we show the level of agreement within the proposed annotations A1, A2, and A3 and compare each of them to the GT. 

In Table~\ref{tab:As_to_gt}, we present a comparison of each annotation A1, A2, and A3 and the gold segmentation reference. We can see that annotation A2 has the highest SEG score, i.e., it is the closest to GT. Annotation A3 has the lowest similarity score. 

In Table~\ref{tab:IAV}, we present the SEG scores between annotations of each annotator (A1, A2, A3) and their majority-fusion result (MV). The table shows how similar MV is to every annotation A1, A2, and A3. The result is more similar to the annotations A2 and A3. The final inter-annotator variability of the proposed annotations for sequence S01 is $0.816 \pm 0.055$. For comparison, the inter-annotator variability for GT is $0.742 \pm 0.048$.

\begin{table}[htb]
\centering
\begin{tabular}{l|l|l|l|l|l|}
\toprule
\multirow{2}{*}{}
        &   \multicolumn{2}{c}{A1}
                &   \multicolumn{1}{c}{A2} 
                        &   \multicolumn{1}{c}{A3} \\
    \cmidrule(lr){2-3}\cmidrule(lr){4-4}\cmidrule(lr){5-5}
                &   S01 &   S02     &   S01   &   S01   \\
    \midrule
\#2d GT masks  &   121  &   103      &   121      &   121      \\
\#matched masks&   119  &   98       &   117       &   106      \\
GT (SEG)       &   0.693  &   0.688   &   \textbf{0.736}    &   0.664      \\
GT (HD)        &   5.08  &  6.54    &   \textbf{4.52}    &   6.10  \\
    \bottomrule
\end{tabular}
  \caption{Quantitative comparison of annotations made by different annotators with the gold segmentation truth annotations. In the table, a Jaccard index-based measure (SEG) and averaged Hausdorff distance (HD) were computed between respective 2D slices for (GT, A1), (GT, A2) and (GT, A3) pairs. See section \ref{sec:Dataset_overview} for more details on SEG measure and matching procedure.}
\label{tab:As_to_gt}
\end{table}

\begin{table}[htb]
\centering

\begin{tabular}{l|l|l|l|l|l|l|}
\multirow{2}{*}{IAV = $0.816 \pm 0.055$}
        &   \multicolumn{1}{c}{A1}
                &   \multicolumn{1}{c}{A2} 
                        &   \multicolumn{1}{c}{A3} \\
    \cmidrule(lr){2-2}\cmidrule(lr){3-3}\cmidrule(lr){4-4}
               &   S01      &   S01 &   S01  \\
    \midrule
FA (SEG)       &     0.738      &   0.857  &   0.857   \\
    \bottomrule
\end{tabular}
  \caption{The table shows the SEG scores between annotations of each annotator (A1, A2, A3) and their majority-fusion result (MV). Thus, the inter-annotator variability for FA of sequence S01 is  $0.816 \pm 0.055$.}
\label{tab:IAV}
\end{table}

\section{Discussion and future work}

Several conclusions can be made based on the quantitative results presented in Section~\ref{sec:Dataset_overview}. First, a consensus of annotations (MV) shed a much better performance even though annotations A1 and A3 exhibit lower similarity to the GT than A2. The lower SEG score for the A1 follows from the over-annotation, i.e., the masks are larger than the masks obtained from other annotators. We can see from Table \ref{tab:all_stats} that A1 is characterized by the largest size of the bounding box and volume. The low score of A3 can be explained by not precise under-segmentation, namely, some tracking markers were not fully redrawn. 
Second, all provided annotations have a perfect one-to-one relation with the tracking truth (TRA), which does not hold for ST. 

As the majority voting has a much better performance than one annotation, in the following month we plan to collect at least 2 annotations for the second sequence to have three human annotations for both sequences. We also want to revise obvious errors in A3 to further improve the quality of provided annotations. The final goal is to have three independent human annotations of both time-lapse sequences of the Fluo-C3DL-MDA231 dataset as well as its corresponding MV fusion.

\section{Conclusion}
\label{sec:pagestyle}

In this paper, we presented a comprehensive description and quantitative experiments for the dataset of annotation that was presented in \cite{melnikova2025study}. The dataset includes fully volumetric annotations for two publicly available time-lapse sequences of fluorescence microscopy images of migrating cells with complex shapes. The dataset can be used for training and validation of cell segmentation and tracking algorithms. Additionally, it can be used for the development and testing the fusion algorithms.  

The annotations can be freely downloaded from CBIA website \url{https://cbia.fi.muni.cz/datasets/#isbi25dataset} or by \url{https://datasets.gryf.fi.muni.cz/isbi2025/Fluo-C3DL-MDA231_Full_Annotations.zip}


\newpage
\bibliographystyle{IEEEbib}
\bibliography{strings,refs}

@inproceedings{melnikova2025study,
  title={Study of Shape Fusion Algorithms for 3D Time-Lapse Microscopy},
  author={Melnikova, A. and Matula, P.},
  booktitle={2025 IEEE 22nd International Symposium on Biomedical Imaging (ISBI)},
  pages={1--5},
  year={2025},
  organization={IEEE}
}

@inproceedings{maska2019toward,
  title={Toward robust fully 3D filopodium segmentation and tracking in time-lapse fluorescence microscopy},
  author={Ma{\v{s}}ka, M.  and others},
  booktitle={2019 IEEE International Conference on Image Processing (ICIP)},
  pages={819--823},
  year={2019},
  organization={IEEE}
}

@article{sorokin2018filogen,
  title={FiloGen: a model-based generator of synthetic 3-D time-lapse sequences of single motile cells with growing and branching filopodia},
  author={Sorokin, D. V. and others},
  journal={IEEE Trans. Med. Imaging},
  volume={37},
  number={12},
  pages={2630--2641},
  year={2018},
  publisher={IEEE}
}

@inproceedings{svoboda2011generation,
  title={Generation of 3D digital phantoms of colon tissue},
  author={Svoboda, D. and Homola, O. and Stejskal, S.},
  booktitle={Image Analysis and Recognition: 8th International Conference, ICIAR 2011, Burnaby, BC, Canada, June 22-24, 2011. Proceedings, Part II 8},
  pages={31--39},
  year={2011},
  organization={Springer}
}

@article{svoboda2009generation,
  title={Generation of digital phantoms of cell nuclei and simulation of image formation in 3D image cytometry},
  author={Svoboda, D. and Kozubek, M. and Stejskal, S.},
  journal={Cytometry Part A: The Journal of the International Society for Advancement of Cytometry},
  volume={75},
  number={6},
  pages={494--509},
  year={2009},
  publisher={Wiley Online Library}
}

@article{mavska2014benchmark,
  title={A benchmark for comparison of cell tracking algorithms},
  author={Ma{\v{s}}ka, M.  and others},
  journal={Bioinformatics},
  volume={30},
  number={11},
  pages={1609--1617},
  year={2014},
  publisher={Oxford University Press}
}

@article{ljosa2012annotated,
  title={Annotated high-throughput microscopy image sets for validation.},
  author={Ljosa, V. and Sokolnicki, K. L. and Carpenter, A. E.},
  journal={Nat. Methods},
  volume={9},
  number={7},
  pages={637--637},
  year={2012}
}

@article{Williams2017,
   author = {Williams, E. and others},
   doi = {10.1038/nmeth.4326},
   issn = {1548-7091},
   issue = {8},
   journal = {Nat. Methods},
   month = {8},
   pages = {775-781},
   title = {Image Data Resource: a bioimage data integration and publication platform},
   volume = {14},
   year = {2017},
}

@article{mavska2023cell,
  title={The Cell Tracking Challenge: 10 years of objective benchmarking},
  author={Ma{\v{s}}ka, M.  and others},
  journal={Nat. Methods},
  pages={1--11},
  year={2023},
  publisher={Nature Publishing Group US New York}
}

@article{chang2015shapenet,
  title={Shapenet: An information-rich 3d model repository},
  author={Chang, A. X.  and others},
  journal={arXiv preprint arXiv:1512.03012},
  year={2015}
}

@article{armeni2017joint,
  title={Joint 2d-3d-semantic data for indoor scene understanding},
  author={Armeni, I. and Sax, S. and Zamir, A. R. and Savarese, S.},
  journal={arXiv preprint arXiv:1702.01105},
  year={2017}
}

@article{guo2020deep,
  title={Deep learning for 3d point clouds: A survey},
  author={Guo, Y. and Wang, H. and Hu, Q. and Liu, H. and Liu, L. and Bennamoun, M. },
  journal={IEEE transactions on pattern analysis and machine intelligence},
  volume={43},
  number={12},
  pages={4338--4364},
  year={2020},
  publisher={IEEE}
}

@inproceedings{sun2020scalability,
  title={Scalability in perception for autonomous driving: Waymo open dataset},
  author={Sun, P. and others},
  booktitle={Proceedings of the IEEE/CVF conference on computer vision and pattern recognition},
  pages={2446--2454},
  year={2020}
}

@inproceedings{tan2020toronto,
  title={Toronto-{3D}: A large-scale mobile {L}i{DAR} dataset for semantic segmentation of urban roadways},
  author={Tan, W. and others},
  booktitle={Proceedings of the IEEE/CVF conference on computer vision and pattern recognition workshops},
  pages={202--203},
  year={2020}
}

@article{kittler1998combining,
  title={On combining classifiers},
  author={Kittler, J. and Hatef, M. and Duin, R. P. and Matas, J.},
  journal={IEEE transactions on pattern analysis and machine intelligence},
  volume={20},
  number={3},
  pages={226--239},
  year={1998},
  publisher={IEEE}
}

@article{langerak2010label,
  title={Label fusion in atlas-based segmentation using a selective and iterative method for performance level estimation (SIMPLE)},
  author={Langerak, T. R. and others},
  journal={IEEE Trans. Med. Imaging},
  volume={29},
  number={12},
  pages={2000--2008},
  year={2010},
  publisher={IEEE}
}

@article{warfield2004simultaneous,
  title={Simultaneous truth and performance level estimation ({STAPLE}): an algorithm for the validation of image segmentation},
  author={Warfield, S. K. and Zou, K. H. and Wells, W. M.},
  journal={IEEE Trans. Med. Imaging},
  volume={23},
  number={7},
  pages={903--921},
  year={2004},
  publisher={IEEE}
}

@article{menze2014multimodal,
  title={The multimodal brain tumor image segmentation benchmark ({BRATS})},
  author={Menze, B. H.  and others},
  journal={IEEE Trans. Med. Imaging},
  volume={34},
  number={10},
  pages={1993--2024},
  year={2014},
  publisher={IEEE}
}

@article{clark2013cancer,
  title={The Cancer Imaging Archive ({TCIA}): maintaining and operating a public information repository},
  author={Clark, K. and others},
  journal={J. Digit. Imaging},
  volume={26},
  pages={1045--1057},
  year={2013},
  publisher={Springer}
}

@article{bauer2017numerical,
  title={A numerical framework for Sobolev metrics on the space of curves},
  author={Bauer, M. and Bruveris, M. and Harms, P. and M{\o}ller-Andersen, J.},
  journal={SIAM Journal on Imaging Sciences},
  volume={10},
  number={1},
  pages={47--73},
  year={2017},
  publisher={SIAM}
}

@article{ulman2017objective,
  title={An objective comparison of cell-tracking algorithms},
  author={Ulman, V. and others},
  journal={Nat. Methods},
  volume={14},
  number={12},
  pages={1141},
  year={2017},
  publisher={Nature Publishing Group}
}

@article{prasad2019cell,
  title={Cell form and function: interpreting and controlling the shape of adherent cells},
  author={Prasad, A. and Alizadeh, E.},
  journal={Trends Biotechnol.},
  volume={37},
  number={4},
  pages={347--357},
  year={2019},
  publisher={Elsevier}
}

@article{alizadeh2020cellular,
  title={Cellular morphological features are predictive markers of cancer cell state},
  author={Alizadeh, E. and others},
  journal={Comput. Biol. Med.},
  volume={126},
  pages={104044},
  year={2020},
  publisher={Elsevier}
}

@article{armato2011lung,
  title={The lung image database consortium ({LIDC}) and image database resource initiative ({IDRI}): a completed reference database of lung nodules on CT scans},
  author={Armato S. G. and others},
  journal={Med. Phys.},
  volume={38},
  number={2},
  pages={915--931},
  year={2011},
  publisher={Wiley Online Library}
}

\end{document}